\documentclass{article}

\usepackage[preprint]{neurips_2026}

\usepackage[utf8]{inputenc}
\usepackage[T1]{fontenc}
\usepackage{hyperref}
\usepackage{url}
\usepackage{booktabs}
\usepackage{amsfonts}
\usepackage{amsmath}
\usepackage{nicefrac}
\usepackage{microtype}
\usepackage{graphicx}
\usepackage{xcolor}
\usepackage{placeins}

\workshoptitle{Workshop on Embodied Spatial Reasoning (ESR)}

\title{Agreement versus Softmax under Spatial Disambiguation}

\author{%
Micha{\l} Zakrzewski\textsuperscript{1,3},
Mariia Shpir\textsuperscript{2,3},
Taras Rumezhak\textsuperscript{3},
Volodymyr Karpiv\textsuperscript{3}\\[0.35em]
\textsuperscript{1}AGH University of Krakow, Krakow, Poland\\
\textsuperscript{2}ETH Zurich\\
\textsuperscript{3}SoftServe
}

\begin{document}

\maketitle

\begin{abstract}
ScanRefer Multiple queries require spatial disambiguation: several
instances of the target class are present, and the referencer must
pick one.
When per-query logits were never logged, a common substitute is
cross-epoch agreement.
On the full ScanRefer validation set ($9{,}508$ queries;
GeoClassifier with ground-truth proposals), object-level softmax is
similarly miscalibrated on Unique and Multiple (ECE $0.068$ vs.\
$0.102$; the scene-bootstrap gap interval includes zero).
Fixed-final agreement on the same checkpoints reports ECE $0.041$ vs.\
$0.272$ (gap $+0.231$).
Modal-vote agreement is $0.045$ vs.\ $0.318$.
Most Multiple errors are same-class wrong-instance ($45.6\%$);
geometric miss (H2) is $59.6\%$, and wrong-category (H1) is $16.7\%$.
A non-monotone H1 pattern does not appear under the default
fixed-final protocol, and we retract it.
Logit-free thresholds rank spatially hard queries poorly.
If only agreement was logged, calibrate abstention on Multiple and
report wrong-instance rates with H1.
\end{abstract}

\section{Introduction}
\label{sec:intro}

Embodied 3D assistants are asked to ground a referring phrase to one
object in a reconstructed scene~\citep{chen2020_scanrefer,achlioptas2020_referit3d,huang2024_chatscene}.
ScanRefer's official Unique/Multiple split is a spatial test.
A Unique query has a single instance of the target class in the scene.
A Multiple query has two or more, so the model has to use geometry
and spatial language (``the chair by the window'') rather than class
identity.
Multiple is the bulk of the benchmark: $7{,}749$ of $9{,}508$
validation queries ($81\%$) over $141$ ScanNet
scenes~\citep{dai2017_scannet}.

Grounding papers report Acc@$0.25$ and Acc@$0.5$.
A deployment that abstains when the model is unsure
needs a confidence score~\citep{geifman2017_selective,guo2017_calibration}.
Identifier-token 3D-LLMs often decode by sampling and discard the
object-level distribution~\citep{huang2024_chatscene,hong2023_3dllm}.
When logits are missing, a common practice is to treat epoch dumps as
votes and use cross-epoch agreement as confidence, as in checkpoint
ensembles~\citep{lakshminarayanan2017_deepensembles}.
That proxy can fabricate a Unique/Multiple calibration gap that
max-softmax on the same checkpoints does not show.
On a GeoClassifier with an object classification head, we recover
native max-softmax on every validation query and compare it to
agreement on the same dumps.
Softmax ECE is $0.068$ on Unique and $0.102$ on Multiple.
Fixed-final agreement, which we take as the default protocol, is
$0.041$ vs.\ $0.272$.
Mean softmax falls with difficulty ($0.771$ Unique vs.\ $0.480$
Multiple), consistent with accuracy ($0.837$ vs.\ $0.377$).
Agreement ECE is a poor spatial-reasoning reliability diagnostic.

Wrong-category hallucination (H1) is secondary on Multiple.
Same-class wrong-instance is $45.6\%$ of Multiple queries; H1 is
$16.7\%$.
Existence-hallucination benches such as 3D-POPE ask whether an
object class is present~\citep{yang2025_3dgrand,li2023_pope}.
On Multiple the class is usually correct and the instance is wrong.

\section{Related work}
\label{sec:related}

ScanRefer~\citep{chen2020_scanrefer} and
ReferIt3D~\citep{achlioptas2020_referit3d} are the standard 3D
referring benchmarks.
Published 3D-LLM grounders are scored by IoU accuracy, not by
calibration or selective
prediction~\citep{huang2024_chatscene,hong2023_3dllm}.

Object hallucination in 2D VLMs is often probed with yes/no
existence questions~\citep{li2023_pope}.
3D-GRAND introduces 3D-POPE, the same existence protocol on 3D
scenes~\citep{yang2025_3dgrand}.
3D-VCD then reduces those existence errors at decode time by
contrasting a scene graph against geometric and semantic
corruptions~\citep{ogunleye2026_3dvcd}.
Those tools measure whether a system invents an absent category.
They do not test whether confidence tracks \emph{which} same-class box
was chosen when several candidates are visible.
We measure that gap on ScanRefer.

Modern networks are often
overconfident~\citep{guo2017_calibration,naeini2015_ece}.
Selective prediction and LLM self-knowledge work assume a usable
score~\citep{geifman2017_selective,hendrycks2017_baseline,kadavath2022_knowsknow}.
When logits are missing, disagreement among models is a standard
stand-in~\citep{lakshminarayanan2017_deepensembles}.
We test whether that stand-in, from one training trajectory, is safe
to threshold on ScanRefer Multiple.

\section{Setup}
\label{sec:setup}

\paragraph{Model.}
The audited system is a GeoClassifier: a geometric 3D referencer
with a frozen DistilBERT query encoder and a classification head over
object proposals, distinct from Chat-Scene's identifier-token LLM.
Proposals are ScanNet ground-truth boxes.
There is no Mask3D detector in this run, so detector error is not
mixed into Unique/Multiple ECE.
The head emits one logit per proposal.
Native confidence is the maximum softmax probability over those
logits.
Prediction is $\arg\max$ of the same vector.
The checkpoint is the 10-epoch \texttt{best\_pos8} run (final dump
parity $9{,}508/9{,}508$ against a fresh softmax pass).

\paragraph{Data.}
ScanRefer validation: $9{,}508$ queries, $141$ scenes, official
Unique/Multiple lookup ($1{,}759$ Unique, $7{,}749$ Multiple).
A query is Unique iff the target ScanRefer class has one instance in
the scene.
Accuracy is identifier match against the GT target.
With GT boxes that match is equivalent to selecting the labelled
instance.

\paragraph{Confidence protocols.}
$E{=}10$ epoch dumps.
\emph{Fixed-final} (default): prediction is the final epoch;
confidence is the fraction of the other $9$ dumps that agree with
that prediction.
\emph{Modal-vote}: prediction is the mode over all $10$ dumps;
confidence is the modal frequency.
Softmax uses the final checkpoint only.
Fixed-final and softmax share every prediction; they differ only in
the score attached to it.
Modal-vote can change the identifier.

\paragraph{Metrics.}
ECE uses $B{=}10$ equal-width bins~\citep{naeini2015_ece}.
Intervals are scene-level bootstrap $95\%$ intervals ($1{,}000$
resamples of $141$ scenes), not query-level draws.
H1 is wrong ScanNet class (``other'' excluded).
Wrong-instance is same-class identifier $\neq$ target.
H2 is box IoU ${<}\,0.05$ against the target.
Abstention at threshold $\tau$ keeps queries with confidence
$\ge \tau$.

\paragraph{Scope.}
One GeoClassifier family, ScanRefer val and Nr3D val (ReferIt3D
natural split~\citep{achlioptas2020_referit3d}), GT proposals.
ScanRefer numbers below are the primary setup; Nr3D is a same-protocol
replication (Section~\ref{sec:nr3d}).
Two other GeoClassifier variants with $E{=}4$ dumps are a coarse
check only (Appendix~\ref{app:variants}).

\section{Results}
\label{sec:results}

\subsection{Softmax versus agreement}
\label{sec:ece}

Table~\ref{tab:lead} is the lead comparison, on the full validation
set.
Softmax ECE is $0.068$ on Unique and $0.102$ on Multiple
(scene-bootstrap $[0.053, 0.097]$ and $[0.085, 0.122]$).
The paired Unique/Multiple gap is $+0.034$
($[-0.002, +0.055]$; $97\%$ of scene resamples positive).
Fixed-final ECE is $0.041$ vs.\ $0.272$ (gap $+0.231$;
$[0.029, 0.070]$ vs.\ $[0.256, 0.289]$).
Modal-vote ECE is $0.045$ vs.\ $0.318$ (gap $+0.273$).
The agreement gap is an order of magnitude larger than the softmax
gap.
Amplification (fixed-final gap minus softmax gap) is $+0.197$
($[0.155, 0.238]$).

\begin{table}[t]
  \centering
  \small
  \caption{Full ScanRefer val ECE. Softmax and fixed-final share the
    final-checkpoint prediction. Modal-vote uses a different
    identifier. Gaps are Multiple minus Unique.}
  \label{tab:lead}
  \setlength{\tabcolsep}{4.5pt}
  \begin{tabular}{@{}lcccc@{}}
    \toprule
    Split & $n$ & Softmax ECE & Fixed-final ECE & Modal ECE \\
    \midrule
    Unique   & $1{,}759$ & $0.068$ & $0.041$ & $0.045$ \\
    Multiple & $7{,}749$ & $0.102$ & $0.272$ & $0.318$ \\
    Gap (M$-$U) &         & $+0.034$ & $+0.231$ & $+0.273$ \\
    \bottomrule
  \end{tabular}
\end{table}

\begin{figure}[t]
  \centering
  \includegraphics[width=\linewidth]{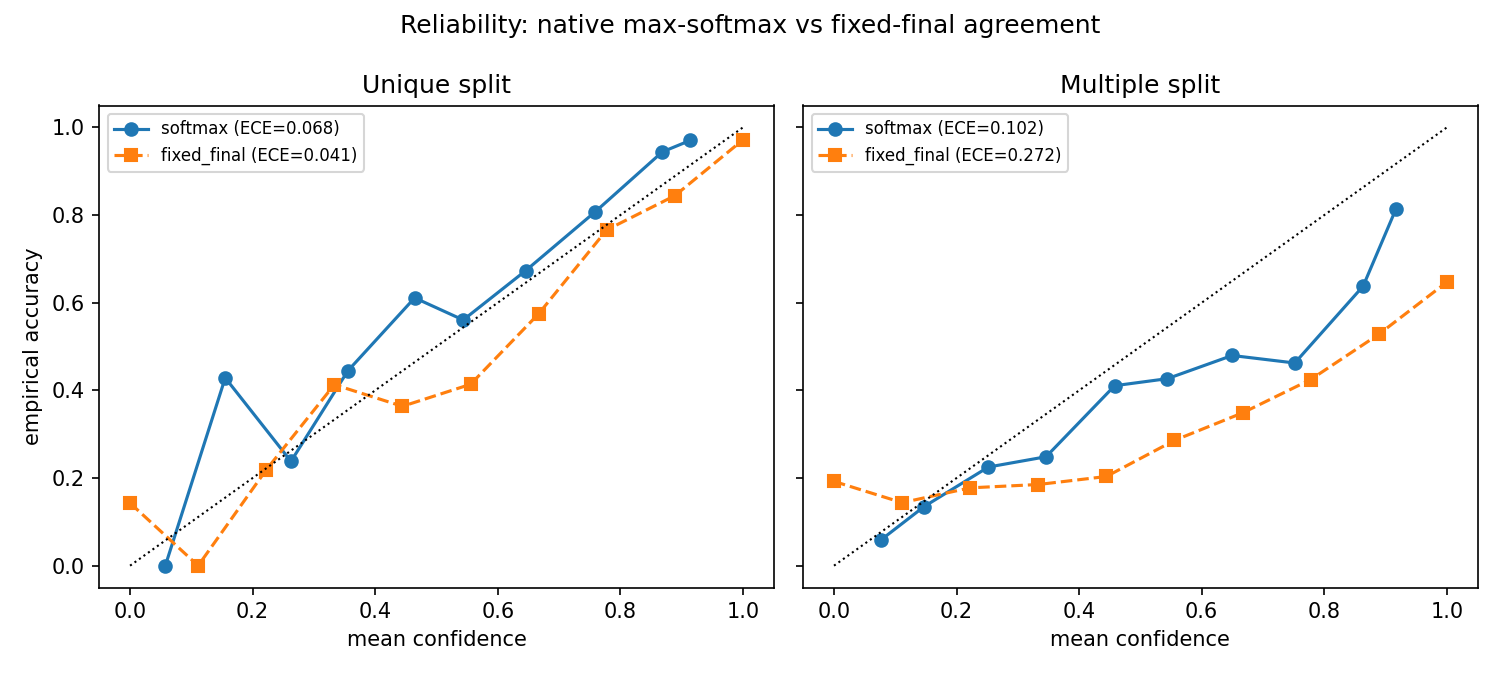}
  \caption{Reliability on Unique and Multiple for native max-softmax
    and fixed-final agreement (same $9{,}508$ queries as
    Table~\ref{tab:lead}). The dotted line is perfect calibration.}
  \label{fig:reliability}
\end{figure}

Figure~\ref{fig:reliability} plots the reliability curves.
On Unique, both scores sit near the diagonal at high confidence.
On Multiple, softmax is overconfident but still tracks accuracy
upward; fixed-final agreement sits well below the diagonal once
confidence exceeds about $0.4$.
Mean softmax ($0.771$ Unique, $0.480$ Multiple) moves with accuracy.
Mean fixed-final agreement does not ($0.873$ vs.\ $0.643$) relative
to the accuracy drop of $0.46$.

\subsection{Wrong-instance errors on Multiple}
\label{sec:errors}

Table~\ref{tab:errors} uses the fixed-final (equivalently softmax)
identifier.
On Multiple, $45.6\%$ of queries are same-class wrong-instance,
$16.7\%$ are H1, and $59.6\%$ are H2.
On Unique, wrong-instance is $1.4\%$ (one GT box of that class), H1
is $14.9\%$, and H2 is $16.0\%$.
Of $5{,}112$ errors overall, $4{,}903$ ($95.9\%$) are H2 misses;
$3{,}531$ of $3{,}555$ wrong-instance errors fall on Multiple.

H1 is almost the same on Unique and Multiple ($14.9\%$ vs.\ $16.7\%$),
so it does not explain the spatial-disambiguation miss.
Wrong-instance errors (wrong chair, cabinet, or door) account for
most of the Multiple drop.

\begin{table}[t]
  \centering
  \small
  \caption{Error types on the final-checkpoint identifier
    (fixed-final / softmax). Rates are fractions of queries, not of
    errors.}
  \label{tab:errors}
  \setlength{\tabcolsep}{5pt}
  \begin{tabular}{@{}lcccc@{}}
    \toprule
    Split & Acc & H1 & Wrong-instance & H2 \\
    \midrule
    Unique   & $0.837$ & $14.9\%$ & $1.4\%$ & $16.0\%$ \\
    Multiple & $0.377$ & $16.7\%$ & $45.6\%$ & $59.6\%$ \\
    \bottomrule
  \end{tabular}
\end{table}

On Multiple, both H2 and wrong-instance fall as fixed-final
agreement rises: H2 $0.80$, $0.70$, $0.49$, $0.32$ and
wrong-instance $0.61$, $0.51$, $0.40$, $0.27$ across buckets
$[0.00,0.40)$, $[0.40,0.70)$, $[0.70,0.90)$, $[0.90,1.00]$.
H1 on those same buckets is $21.5\%$, $21.5\%$, $12.2\%$, $8.7\%$.
That sequence is flat, then decreasing.
A medium-confidence H1 peak reported under modal-vote does not
appear here, so we retract non-monotone H1 as a default-protocol
claim.

\subsection{Abstention and ranking on Multiple}
\label{sec:abstain}

Table~\ref{tab:abstain} is the Multiple abstention curve under
fixed-final agreement.
At full coverage, accuracy is $0.377$, wrong-instance $0.456$, H2
$0.596$.
Keeping only agreement ${=}\,1$ (coverage $0.184$) raises accuracy
to $0.647$ and cuts wrong-instance to $0.266$ and H2 to $0.319$.
Agreement still ranks some errors as certain: among Multiple queries
with agreement $1$, the error rate is $35.3\%$.
A softmax threshold with the same coverage ($18.4\%$;
$\tau{\approx}\,0.821$) has error $29.4\%$.

\begin{table}[t]
  \centering
  \small
  \caption{Fixed-final abstention on Multiple ($n{=}7{,}749$).
    $\tau$ is agreement with the final checkpoint among the other
    nine dumps.}
  \label{tab:abstain}
  \setlength{\tabcolsep}{4pt}
  \begin{tabular}{@{}cccccc@{}}
    \toprule
    $\tau$ & Coverage & Acc & Wrong-inst. & H1 & H2 \\
    \midrule
    $0$     & $1.000$ & $0.377$ & $0.456$ & $0.167$ & $0.596$ \\
    $5/9$   & $0.671$ & $0.472$ & $0.387$ & $0.141$ & $0.502$ \\
    $8/9$   & $0.325$ & $0.595$ & $0.316$ & $0.089$ & $0.371$ \\
    $1$     & $0.184$ & $0.647$ & $0.266$ & $0.087$ & $0.319$ \\
    \bottomrule
  \end{tabular}
\end{table}

For selective prediction, score ${=}\,1$ minus confidence.
On Multiple, softmax AUROC is $0.747$ and AURC is $0.432$;
fixed-final AUROC is $0.709$ and AURC is $0.469$.
On Unique the ranking order flips (fixed-final AUROC $0.868$ vs.\
softmax $0.843$), which is the easy split with $19\%$ of the mass.
A threshold tuned on pooled agreement will spend most of its
coverage budget on Unique, where agreement is already high
($60\%$ of Unique queries have agreement $1$).

If a global mapping is fitted to pooled agreement and applied to
both splits, Unique ECE gets worse.
Scene-grouped out-of-fold Platt scaling~\citep{platt1999_scaling} of
raw fixed-final agreement yields Unique ECE $0.222$ (vs.\ raw
$0.041$) and Multiple ECE $0.068$ (vs.\ $0.272$).
The mapping drags Unique confidence toward the pooled accuracy.
Split-specific Platt keeps Unique ECE $0.030$ and Multiple $0.023$.
A single temperature on agreement trades Unique ECE for Multiple ECE.
The Multiple-conditioned rule keeps both in range.

\subsection{Candidate multiplicity}
\label{sec:mult}

\begin{figure}[t]
  \centering
  \includegraphics[width=\linewidth]{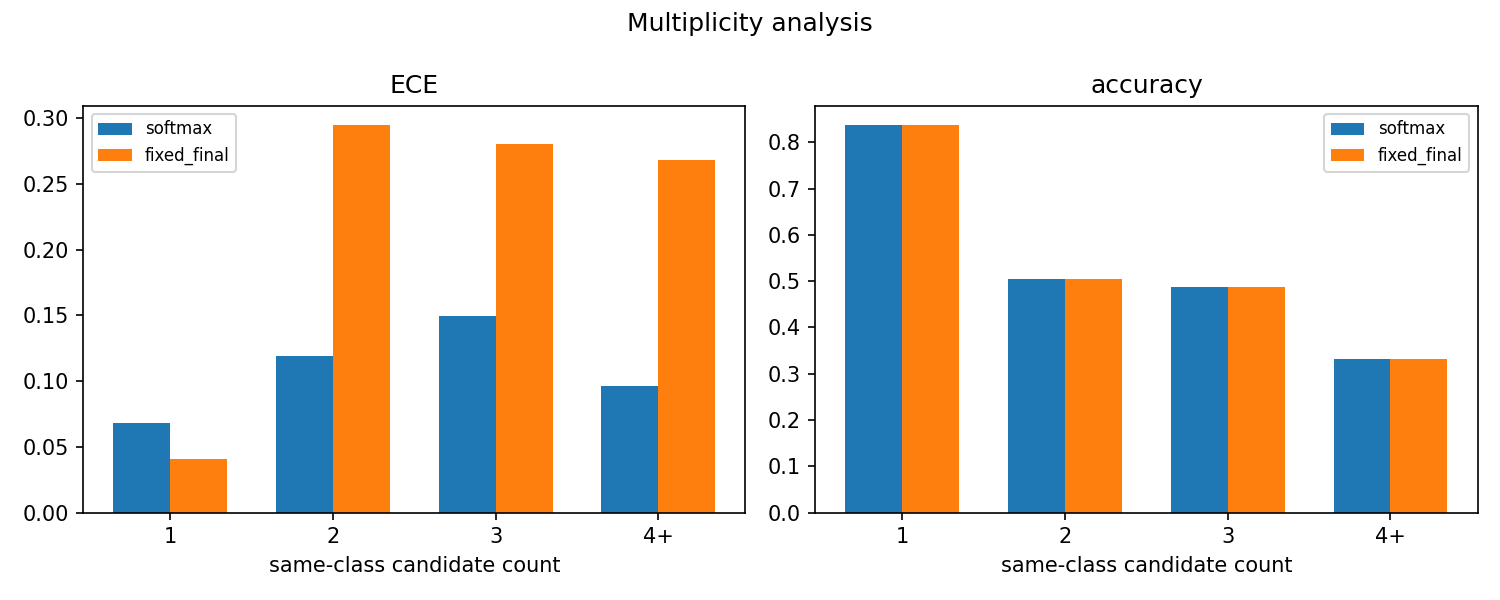}
  \caption{ECE and accuracy by same-class candidate count on
    ScanRefer val. Count $1$ is the Unique split. Counts $2$ and
    above are Multiple.}
  \label{fig:mult}
\end{figure}

Figure~\ref{fig:mult} bins queries by the number of same-class GT
boxes (the official Unique rule is count ${=}\,1$, matched on
$9{,}508/9{,}508$ rows).
Softmax ECE is $0.068$, $0.119$, $0.149$, $0.096$ at counts $1$, $2$,
$3$, $4{+}$.
Fixed-final ECE is $0.041$, $0.295$, $0.281$, $0.268$.
Once a second same-class candidate appears, agreement ECE stays
high.
Accuracy falls from $0.837$ at count $1$ to $0.505$, $0.487$,
$0.332$ at $2$, $3$, $4{+}$; mean fixed-final confidence only goes
from $0.873$ to $0.792$, $0.746$, $0.593$.
Softmax mean confidence tracks that drop more closely ($0.771$ to
$0.622$, $0.608$, $0.428$).
From count $2$ onward, agreement ECE stays high while accuracy falls.

\subsection{Nr3D replication}
\label{sec:nr3d}

We repeat the audit on Nr3D validation ($8{,}584$ queries, $130$
scenes) with the same GeoClassifier, $E{=}10$ dumps, and GT
proposals.
Every Nr3D stimulus lists at least one same-class distractor, so
there is no count-${=}\,1$ tier.
We therefore compare easy (count ${=}\,2$; $n{=}4{,}118$) to hard
(count $\ge 3$; $n{=}4{,}466$), the dataset-native multiplicity split.

Table~\ref{tab:nr3d} shows the same qualitative pattern as
Table~\ref{tab:lead}.
Softmax ECE moves little between easy and hard ($0.177$ vs.\
$0.195$; paired scene gap $+0.018$, bootstrap $[-0.020, +0.051]$).
Fixed-final ECE is higher on both tiers ($0.279$ vs.\ $0.364$) and
the gap is larger ($+0.083$; $[0.046, 0.115]$).
Amplification (fixed-final gap minus softmax gap) is $+0.065$
($[0.033, 0.096]$), about one third of the ScanRefer magnitude.
Overall, fixed-final ECE is $0.323$ vs.\ softmax $0.186$.
Same-class wrong-instance is $64.9\%$ of Nr3D errors ($3{,}912$ of
$6{,}025$).

\begin{table}[t]
  \centering
  \small
  \caption{Nr3D val ECE by multiplicity tier (easy $=$ count $2$,
    hard $=$ count $\ge 3$). Gaps are hard minus easy.}
  \label{tab:nr3d}
  \setlength{\tabcolsep}{4.5pt}
  \begin{tabular}{@{}lcccc@{}}
    \toprule
    Tier & $n$ & Softmax ECE & Fixed-final ECE \\
    \midrule
    Easy (count $2$)   & $4{,}118$ & $0.177$ & $0.279$ \\
    Hard (count $\ge 3$) & $4{,}466$ & $0.195$ & $0.364$ \\
    Gap (H$-$E)        &           & $+0.018$ & $+0.083$ \\
    \bottomrule
  \end{tabular}
\end{table}

\section{Limitations and deployment}
\label{sec:limits}

The model is a GeoClassifier with GT boxes, not a Chat-Scene-class
LLM with Mask3D proposals.
Identifier softmax is available here because the head is an object
classifier.
An LLM that only emits \texttt{<OBJ>} tokens still needs
instrumented decoding before the same softmax comparison can be run.
We did not retrain under detected boxes, so detector false
positives and missed instances are out of scope.
ScanRefer is single-target; multi-object referring is untested.
Fixed-final uses one training trajectory.
$E{=}4$ variants (Appendix~\ref{app:variants}) reproduce the
amplification only as a coarse qualitative check.

\paragraph{Deployment.}
Log the object-level softmax (or the identifier-token distribution)
and threshold that.
If the stored trace is epoch identifiers only, treat agreement as a
ranking score that is miscalibrated on Multiple: set $\tau$ on
Multiple, not on Unique or on the pool, and report wrong-instance
and H2 with H1.
Low Unique agreement ECE does not establish reliability when only one
instance of the class appears in the scene.

\FloatBarrier
\section{Conclusion}
\label{sec:conclusion}

On ScanRefer Multiple, the spatial disambiguation regime, object-level
softmax is similarly miscalibrated on Unique and Multiple (ECE
$0.068$ vs.\ $0.102$).
Cross-epoch agreement reports a large gap that softmax does not
show ($+0.231$ fixed-final, $+0.273$ modal-vote).
The same amplification sign replicates on Nr3D ($+0.065$), with no
Unique tier because every query has a same-class distractor.
Agreement-based abstention gives poor selective-prediction ranking on
Multiple.
Wrong-instance is the largest Multiple error category ($45.6\%$), with
H2 at $59.6\%$; H1 is $16.7\%$.
For deployed 3D referencers, log object-level softmax when available.
If only epoch identifiers are stored, use Multiple-specific agreement
thresholds and report wrong-instance and H2 rates alongside H1.

\bibliographystyle{plainnat}
\bibliography{references}

\appendix

\section{Coarse variant check}
\label{app:variants}

Two other GeoClassifier runs on the same ScanRefer val set and GT
proposal files, each with four epoch dumps, show fixed-final
Unique/Multiple ECE gaps of $+0.159$ and $+0.169$, while their
softmax gaps are slightly negative.
The four-level agreement grid $\{0, 1/3, 2/3, 1\}$ is too coarse
for a quantitative ECE claim; the point is only that the
amplification is not unique to \texttt{best\_pos8}.

\end{document}